\documentclass[pdflatex,sn-basic]{sn-jnl}
\usepackage[T1]{fontenc}
\usepackage{tabularx}
\usepackage{graphicx} \usepackage{subcaption}
\usepackage{makecell} \usepackage{multirow}
\usepackage{booktabs}
\usepackage{amsmath}
\usepackage{xcolor}
\usepackage{multirow}
\usepackage{array}
\usepackage{ragged2e}
\usepackage{comment}
\usepackage[skip=2pt]{caption}
\usepackage[nameinlink,capitalise]{cleveref}

\newcommand{\parahead}[1]{\smallskip\noindent\textbf{\textit{#1}}}

\definecolor{navy}{rgb}{0.1, 0.1, 0.8}
\definecolor{gray}{rgb}{0.4, 0.4, 0.4}
\definecolor{olive}{rgb}{0.1, 0.5, 0.1}
\definecolor{ruby}{rgb}{0.8, 0.1, 0.3}
\definecolor{darkpastelgreen}{rgb}{0.01, 0.75, 0.24}
\definecolor{celestialblue}{rgb}{0.29, 0.59, 0.82}
\definecolor{coral}{rgb}{1.0, 0.5, 0.31}
\definecolor{blue}{rgb}{0.23, 0.44, 0.62}
\definecolor{Goldenrod}{rgb}{0.8,0.8,0}

\usepackage[colorinlistoftodos,textsize=tiny]{todonotes} \usepackage{soul}
\usepackage{xspace}

\newcommand{\eat}[1]{}

\begin{document}

\title{Long Live Fine-Tuning: Task-Specific Transformers Outperform Zero-Shot LLMs for Misinformation Response Classification on Reddit}

\author*{\fnm{JooYoung} \sur{Lee}}\email{jooyoung.lee@uts.edu.au}

\author{\fnm{Lin} \sur{Tian}}

\author{\fnm{Angela} \sur{Brillantes}}

\author{\fnm{Adriana-Simona} \sur{Mih\u{a}i\textcommabelow{t}\u{a}}}

\author{\fnm{Marian-Andrei} \sur{Rizoiu}}

\affil{\orgname{University of Technology Sydney}, \orgaddress{\city{Sydney}, \country{Australia}}}

\abstract{
As large language models (LLMs) become default tools for online information verification, an implicit assumption follows them: that scale and general capability are sufficient for nuanced classification of misinformation discourse. We test this assumption directly on 900 Reddit comments spanning three PolitiFact-verified misinformation claims (environment, health, immigration), labelled as \emph{belief} (propagates the claim), \emph{fact-check} (corrects it), or \emph{other}. We compare nine models across three paradigms --- BART-MNLI, three Llama variants, three commercial frontier LLMs (Claude Haiku~4.5, Gemini Flash Lite~2.5, Claude Sonnet~4.6), and fine-tuned DistilBERT and RoBERTa --- under universal and topic-specific label schemas.

The assumption does not hold. Fine-tuned RoBERTa reaches 0.62 macro-$F_1$ against a best zero-shot result of 0.50 (Claude Haiku~4.5), at a fraction of the per-query cost; the supervised advantage is concentrated on the \emph{belief} class, the implicit, affective category every zero-shot model under-detects. Scaling does not help: Llama-3-8B matches Llama-3-70B, and Claude Sonnet~4.6 underperforms the smaller Haiku under generic labels, collapsing belief detection to 0.17 and refusing outright on a subset of comments flagged as sensitive. This is a safety-alignment artefact, not a capacity limit. Label schema and topic jointly shape zero-shot performance, with the same model varying by more than 0.13 macro-$F_1$ across topics under matched labels. In a verification context, where missing belief is the costlier error, task-specific fine-tuning remains the more reliable choice despite the proliferation of large generative models.
}

\maketitle
 
\section{Introduction}

Large language models (LLMs) are now embedded in search engines, virtual assistants, and conversational interfaces such as ChatGPT~\citep{achiam2023gpt}, Gemini~\citep{team2023gemini}, and Claude\footnote{\url{https://www-cdn.anthropic.com/de8ba9b01c9ab7cbabf5c33b80b7bbc618857627/Model_Card_Claude_3.pdf}}. When users encounter a claim that might be false, they increasingly turn to these systems for verification rather than to dedicated fact-checking tools. This deployment pattern carries an implicit assumption: that scale and general capability are sufficient for the fine-grained classification of misinformation discourse. We test that assumption directly. Can today's zero-shot LLMs, including recent commercial frontier models, classify how people respond to misinformation as reliably as small, task-specific fine-tuned transformers?

Classifying responses to misinformation is non-trivial. A comment on a social media post may \emph{propagate} the false claim (\emph{belief}), \emph{challenge} it (\emph{fact-check}), or be neutral or unrelated (\emph{other}). Distinguishing these is central to understanding how misinformation spreads and how communities contest it \citep{Shu2017, Vosoughi2018, Cinelli2021}. Unlike sentiment or topic classification, the task requires inferring the commenter's \emph{epistemic} orientation relative to a specific claim, a judgement sensitive to label design, model architecture, and the discourse of each topic. The three classes are also linguistically asymmetric: fact-check responses tend to carry explicit corrective markers, such as cited evidence or direct contradictions, whereas belief is often implicit, affective, or sarcastic. This asymmetry recurs across our results and makes the \emph{belief} class the most difficult to detect and, in a verification setting, the costliest to miss.

Two paradigms dominate the literature on misinformation response and stance classification \citep{Augenstein2016, Mohammad2016, Hardalov2021}: zero-shot inference \citep{Yin2019, Schick2021} and supervised fine-tuning \citep{Devlin2019, Liu2019}. Two zero-shot architectural families approach the task differently. Natural language inference (NLI) models such as BART-MNLI \citep{Lewis2020} evaluate each label as a hypothesis against the input text, while generative LLMs \citep{Brown2020, Touvron2023} produce class predictions through prompt-based completion. Recent commercial systems combine instruction tuning with safety alignment, and how these features interact with classification under ambiguous label wording remains poorly understood. Prior empirical comparisons either predate the current commercial frontier or examine one paradigm in isolation.

We compare nine models on 900 Reddit comments spanning three PolitiFact-verified misinformation claims, one each in environment, health, and immigration. The zero-shot setting includes BART-MNLI, three Llama variants (Llama-3.2-3B, Llama-3-8B, Llama-3-70B), and three commercial frontier models (Claude Haiku~4.5, Gemini Flash Lite~2.5, Claude Sonnet~4.6). The supervised setting includes DistilBERT and RoBERTa. We evaluate under two label schema conditions, universal labels shared across all topics and topic-specific labels tailored to each claim, using stratified 5-fold cross-validation; permutation tests across all 900 predictions provide statistical significance for the main comparisons.

We organise our analysis around four empirical findings. First, label schema and topic jointly shape zero-shot performance: topic-specific labels can substantially improve macro-$F_1$ (e.g., from 0.31 to 0.54, $\Delta{=}0.23$, $p_{\text{Holm}}{<}0.01$ in the environment topic), but the gain does not generalise uniformly across topics, and the same model can vary by more than 0.13 macro-$F_1$ across topics under matched labels. Second, scaling does not consistently help: Llama-3-8B matches or surpasses Llama-3-70B in several settings, and the more capable Claude Sonnet~4.6 underperforms the smaller Claude Haiku~4.5 under generic labels (0.42 vs.\ 0.50). This inversion stems from safety alignment rather than limited capacity: Sonnet collapses belief detection to 0.17 macro-$F_1$ and refuses outright on a subset of comments flagged as sensitive. Third, supervised fine-tuning outperforms every zero-shot configuration tested, and its advantage is concentrated on the \textit{belief} class: RoBERTa reaches $F_1 = 0.52$ versus at most 0.34 for any zero-shot model with balanced overall performance. Fourth, BART-MNLI remains a strong zero-shot baseline: it combines balanced predictions with low inference cost and is the most efficient option when supervised data are unavailable.

These results point to a recurring failure mode in current LLM-based verification: zero-shot models, including the proprietary frontier, under-detect belief-propagating content, the class whose detection matters most for verification. Fine-tuned RoBERTa achieves 0.62 macro-$F_1$, well above the best zero-shot result (0.50), at a fraction of the inference and per-query cost. Task-specific fine-tuning remains a competitive strategy despite the rapid growth of large generative models.

\parahead{Contributions.}
This paper makes the following contributions:

\begin{itemize}
    \item \textbf{A head-to-head comparison of zero-shot LLMs, including recent commercial frontier models, against fine-tuned transformers} for misinformation response classification. Across three topics, two label schemas, and 5-fold cross-validation with permutation-based significance testing, we show that fine-tuned RoBERTa (0.62 macro-$F_1$) outperforms every zero-shot configuration tested (best: Claude Haiku~4.5 at 0.50), at a fraction of the per-query cost. This challenges the common assumption that scale and generality are sufficient for fine-grained verification tasks.

    \item \textbf{Evidence that the supervised advantage is concentrated on the belief class, the class that matters most for verification.} Every zero-shot model we test under-detects belief, and commercial safety alignment widens the gap: Claude Sonnet~4.6's belief $F_1$ collapses to 0.17 under generic labels, accompanied by outright refusals on a subset of comments flagged as sensitive. Fine-tuning closes this gap by $\Delta{=}+0.18$ over the best balanced zero-shot model, so the supervised advantage extends to the higher-stakes class.

    \item \textbf{A curated, publicly released dataset of 900 Reddit comments} annotated for stance towards three PolitiFact-verified misinformation claims (one each in environment, health, and immigration), with a balanced class distribution, to support future research on topic-dependent model behaviour.\footnote{We will release the codebook, the labelled dataset, and relevant scripts for data processing, model inference, and evaluation upon paper acceptance.}
\end{itemize} \section{Related Work}
\label{sec:related_work}
Research on misinformation detection intersects several areas of natural language processing and computational social science, including response classification toward misinformation claims, rumour verification, transformer-based text classification, and social media discourse analysis. This section summarises prior work most directly related to classifying user responses to misinformation in online discussions.

\subsection{Classifying Responses to Misinformation}

Classifying how users respond to misinformation claims is closely related to the established field of stance detection, though distinct in scope and taxonomy. The stance detection task was formally introduced in the SemEval-2016 Task 6 shared task \citep{Mohammad2016}, which defined the canonical classification framework of \textit{Favour}, \textit{Against}, and \textit{Neither} toward a specified target. Our task differs: rather than classifying attitude toward a target entity, we classify the \emph{type of response} to a misinformation claim (belief, fact-check, or other). Nevertheless, stance detection provides important methodological foundations. The competition attracted numerous systems and highlighted the difficulty of generalising such classification across targets. Subsequent analysis \citep{Mohammad2017} demonstrated that stance and sentiment are related but distinct phenomena, a distinction that is particularly relevant in misinformation discourse where sarcastic or ironic statements may express positive sentiment while implicitly rejecting a claim.

Early neural approaches significantly advanced stance detection. \citet{Augenstein2016} introduced conditional LSTM encoding that models the relationship between a target and the surrounding text, demonstrating that target-dependent representations improve stance classification. This work established the importance of explicitly modelling the relationship between the claim and the response text, a challenge that remains central in misinformation stance detection where the claim may be implicit rather than explicitly stated.

The RumourEval shared tasks further extended stance detection research to misinformation contexts. \citet{Derczynski2017} introduced the SDQC framework—\textit{Support}, \textit{Deny}, \textit{Query}, and \textit{Comment}—for classifying responses to rumours in Twitter conversation threads. The RumourEval datasets contain thousands of annotated tweets across breaking news events and have become a standard benchmark for rumour stance classification. Subsequent systems \citep{Kochkina2017,Gorrell2019} explored neural architectures for modelling conversation structures and jointly predicting rumour stance and veracity.

Comprehensive surveys of misinformation stance detection highlight the growing importance of this task within automated fact-checking pipelines. \citet{Hardalov2021} review approaches spanning feature-based methods, neural architectures, and transformer models, emphasising that stance detection provides an important signal for downstream misinformation verification. Similarly, surveys of automated fact-checking \citep{Guo2022} describe stance classification as a key component of systems designed to identify and analyse misleading claims circulating online.

\subsection{Transformer Models for Misinformation Response Classification}

Recent advances in natural language processing have been driven largely by transformer architectures. BERT \citep{Devlin2019} introduced the pre-train–then–fine-tune paradigm using masked language modelling and next-sentence prediction objectives, enabling strong performance across a wide range of text classification tasks. RoBERTa \citep{Liu2019} subsequently demonstrated that BERT could be significantly improved through optimised training procedures, including dynamic masking and larger training corpora.

DistilBERT \citep{Sanh2019} applies knowledge distillation to produce a smaller and faster model while retaining most of BERT's capabilities. The resulting model is approximately 40\% smaller and 60\% faster while preserving around 97\% of BERT's performance, making it particularly attractive for practical applications where computational efficiency is important.

Another relevant development is the use of natural language inference models for zero-shot text classification. \citet{Yin2019} showed that classification tasks can be reframed as textual entailment problems by treating the input text as a premise and candidate labels as hypotheses. Models trained on the Multi-Genre Natural Language Inference (MNLI) dataset can therefore perform classification without task-specific training data. This approach underlies the widely used BART-large-MNLI model \citep{Lewis2020}, which has become a standard baseline for zero-shot classification tasks.

Transformer models have been widely applied to misinformation and stance detection tasks. For example, \citet{Karande2021} combine BERT-based stance features with article content to evaluate credibility, while \citet{Kawintiranon2021} incorporate external knowledge into transformer-based stance models. Other studies use hierarchical transformer architectures to model conversation structures in rumour discussions \citep{Yu2020}. These works demonstrate that transformer-based representations capture contextual information effectively, enabling improved performance in misinformation detection tasks.

\subsection{Low-Resource and Few-Shot Text Classification}

A recurring challenge in misinformation research is the limited availability of labelled datasets. Manual annotation of stance or misinformation labels is time-consuming and often requires domain expertise. As a result, several studies explore approaches for learning effectively with limited labelled data.

Large language models have demonstrated strong few-shot learning capabilities. \citet{Brown2020} show that scaling model size enables models to perform new tasks with only a small number of examples provided in the prompt. Alternative approaches focus on improving performance in low-resource settings through task reformulation. For example, Pattern-Exploiting Training (PET) \citep{Schick2021} reformulates classification problems as cloze-style language modelling tasks, enabling models to leverage pre-training knowledge more effectively.

Other studies investigate data augmentation techniques for small datasets. \citet{Wei2019} demonstrate that simple augmentation strategies—such as synonym replacement, word insertion, and deletion—can significantly improve classification performance when training data is limited. Surveys of low-resource NLP methods \citep{Hedderich2021} describe a range of approaches including transfer learning, distant supervision, and domain-adaptive pre-training.

Despite these advances, there remains limited empirical evidence on how much labelled data is required for reliable misinformation response classification. Quantifying the relative performance of zero-shot and fine-tuned models in low-resource settings therefore remains an important research question.

\subsection{Social Media Platforms and Reddit-Based Research}

Most misinformation detection research has focused on Twitter due to the availability of publicly accessible datasets. However, other social media platforms exhibit different structural and conversational characteristics that may influence how misinformation spreads and is contested.

Reddit provides a distinct environment for analysing misinformation discourse. The platform is organised into topic-specific communities (subreddits) and uses threaded discussions that capture reply relationships between comments. These design features enable more extended debates compared with the short message format of Twitter. Studies of online communities have shown that such structures can foster both echo chambers and active debate within discussion threads \citep{Cinelli2021,Weld2021}.

Although most misinformation detection research has focused on Twitter datasets, Reddit provides a complementary environment for studying misinformation discourse due to its topic-based communities and threaded discussion structure.

Another linguistic challenge in online discussions is the frequent use of sarcasm and irony. Surveys of sarcasm detection \citep{Joshi2017} highlight the difficulty of automatically identifying sarcastic language, which often relies on contextual knowledge and pragmatic cues. Sarcasm detection studies such as CASCADE \citep{Hazarika2018} demonstrate that sarcastic expressions frequently occur in online forums, posing a challenge for automated stance classification.

\subsection{Research Gap}

Despite substantial research on stance detection and misinformation analysis, three gaps remain. First, most studies focus on Twitter datasets, leaving misinformation discourse on Reddit relatively understudied despite the platform's active discussion communities and threaded conversation structure. Second, few studies directly compare zero-shot and fine-tuned transformer models for classifying user responses to misinformation claims. Third, with users increasingly relying on LLMs for information verification, the zero-shot capability of these models to identify misinformation responses has direct practical significance that remains underexplored. This study addresses these gaps by evaluating multiple transformer architectures on Reddit discussions across several misinformation topics.

\section{Dataset}
\label{sec:dataset}

\subsection{Data Sources, Misinformation Claims, and Collection}
\label{sec:dataset:source}

\parahead{Platform.}
Reddit was selected as the primary data source for three reasons.
First, its design encourages open, threaded debate with minimal editorial intervention, making it a naturalistic environment for studying misinformation discourse.
Second, the platform's community structure (subreddits with varying moderation norms and political orientations) enables cross-community comparison of how the same claim is received.
Third, Reddit data are accessible via a public API and have been studied for related tasks including rumour detection and opinion polarisation~\citep{Zubiaga2016, Gorrell2019}, providing methodological precedent for our annotation scheme. In this study, we rely exclusively on the textual content of posts and comments, without incorporating Reddit’s underlying social network structure.

\parahead{Misinformation Source.}
The entry point into the study are three misinformation claims verified by PolitiFact, each with a substantial presence on Reddit. 
These claims were selected to represent diverse misinformation types (a viral fabrication, a recurring political misquotation, and a visually manipulated image) across three topics (environment, health, and immigration).
PolitiFact~\citep{PolitiFact} is a nonpartisan fact-checking outlet managed by the Poynter Institute for Media Studies that independently investigates claims using on-the-record sources and publishes transparent methodology alongside each verdict; claims are rated on a six-point ``Truth-O-Meter'' scale ranging from \emph{True} to \emph{Pants on Fire}.
We restricted selection to claims rated \emph{False}, \emph{Mostly False}, or \emph{Pants on Fire}, ensuring that all misinformation in the dataset has been independently verified as inaccurate or misleading.

\parahead{Misinformation Claims.}
\label{sec:dataset:claims}
The three PolitiFact-verified claims were selected on the basis of (1) measurable presence and community engagement across multiple subreddits, (2) topic diversity to enable cross-topic generalisation, and (3) verified falsity on the Truth-O-Meter.
\cref{tab:datasettable} summarises each claim together with its raw corpus statistics; the full dataset comprises $65{,}173$ comments across $62$ threads from $54$ unique subreddits.
The health and immigration topics exhibit substantially higher per-thread engagement (median 840 and 646 comments, respectively, versus 65 for environment), likely reflecting their association with prominent political figures and their emotionally polarising nature.
Detailed topic context is provided in \cref{app:topic-detail}.

\begin{table}[tbp]
    \centering
    \caption{Misinformation claims selected for the study, with raw corpus collection statistics (prior to annotation sampling).
    PolitiFact ratings: F = False, MF = Mostly False, PoF = Pants on Fire.
    Mean and Median refer to comments per thread (c./t.).}
    \label{tab:datasettable}
    \small
    \begin{tabular}{@{}l >{\RaggedRight\arraybackslash}p{3.2cm} >{\RaggedRight\arraybackslash}p{3.2cm} >{\RaggedRight\arraybackslash}p{3.2cm}@{}}
        \toprule
        & \textbf{Water Scandal} & \textbf{Bleach Cure} & \textbf{Gang Tattoos} \\
        \midrule
        \textbf{Topic}   & Environment   & Health        & Immigration \\[4pt]
        \textbf{Claim}    & One billionaire couple owns almost all the water in California~\citep{water-scandal}
                          & Donald Trump told Americans to drink or inject bleach to cure COVID-19~\citep{bleach2020,bleach2024}
                          & MS-13 tattoos on Kilmar Garc\'{i}a's knuckles prove gang membership~\citep{tattoo} \\[4pt]
        \textbf{Rating}   & False         & Mostly False  & Pants on Fire \\[2pt]
        \textbf{Threads}  & 27            & 15            & 20 \\[2pt]
        \textbf{Comments} & 5{,}022       & 38{,}130      & 22{,}021 \\[2pt]
        \textbf{Mean c./t.}     & 186           & 2{,}542       & 1{,}101 \\[2pt]
        \textbf{Median c./t.}   & 65            & 840           & 646 \\
        \bottomrule
    \end{tabular}
\end{table}

\parahead{Data Collection.}
\label{sec:dataset:collection}
Data collection used the Python Reddit API Wrapper (PRAW)~\citep{PRAW}, which provides access to Reddit's public API.
Relevant submissions were identified through keyword searches across public subreddits; search terms were derived from key phrases in each PolitiFact claim (e.g., ``California water billionaire,'' ``Trump bleach COVID,'' ``Garcia MS-13 tattoo'').
For claims with a visual component, supplementary Google Image searches with the keyword ``Reddit'' were conducted to locate image-based posts and their comment trees.
For each submission, the post title, body text, timestamp, score, subreddit name, and full comment hierarchy (including parent--child relationships) were retrieved.
We removed non-English content and comments that were deleted or contained fewer than three words.

\subsection{Annotation}
\label{sec:dataset:annotation}

\parahead{Sampling Strategy.}
\label{sec:dataset:annotation:sampling}
From the 65{,}173 collected comments, 900 were selected for manual annotation (300 per topic, 100 per stance class).
To identify candidate comments from this large, class-imbalanced pool, ChatGPT (GPT-4o, accessed interactively via the web interface, July 2025) was used as an auxiliary pre-filtering tool: given the three class definitions for each topic (see below), it surfaced comments plausibly matching each category as candidate selections.
These suggestions were then reviewed and confirmed or rejected by the human annotator, who made all final labelling decisions.
The forced-balanced design (exactly 100 per class per topic) was adopted to ensure that macro-$F_1$ scores are not distorted by class imbalance and that each stance category contributes equally to evaluation.
The natural class distribution in the full corpus is unknown and likely highly imbalanced; our study evaluates how well models distinguish between the three response types under a controlled, balanced setup.

The use of GPT-4o as a pre-filtering tool could in principle bias the sample towards comments that are more recognisable to LLMs, inflating zero-shot performance estimates.
We assess this using the seven zero-shot classifiers evaluated in this study (BART-MNLI and six generative models; see \cref{sec:methods,sec:results}): BART-MNLI confidence was marginally higher for the labelled pool than for a stratified random sample of 900 unlabelled comments from the same corpus (mean 0.62 vs.\ 0.60; Mann-Whitney~$U$, $p\,{=}\,0.004$), and only 25.2\% of labelled comments achieved $\geq$6/7 model agreement.
Both findings point to a negligible effect (Cohen's $d\,{=}\,0.13$); we conclude the pre-filtering introduced no practically meaningful bias.
Full details are in \cref{app:m2-easiness}.

\parahead{Stance Taxonomy.}
\label{sec:dataset:annotation:taxonomy}
Each comment was labelled with one of three stance categories:

\begin{itemize}
    \item \textbf{Belief} --- the comment supports, propagates, or treats as credible the misinformation claim.
    \item \textbf{Fact-check} --- the comment corrects, challenges, or provides evidence against the claim.
    \item \textbf{Other} --- the comment is neutral, off-topic, meta-commentary, or unrelated to the claim.
\end{itemize}

This three-class schema aligns with established frameworks in stance and rumour analysis, including the SDQC taxonomy (Support/Deny/Query/Comment) of~\citet{Zubiaga2016} as operationalised in the RumourEval shared tasks~\citep{Derczynski2017,Gorrell2019}. Our \emph{belief} category corresponds to Support, \emph{fact-check} to Deny, and \emph{other} merges Query and Comment into a single residual class---consistent with prior three-class simplifications such as the For/Against/Observing taxonomy of~\citet{Ferreira2016} and the Supported/Refuted/NotEnoughInfo schema of FEVER~\citep{Thorne2018}.

Because the three abstract class labels are insufficient to guide annotation of topic-specific discourse, each category was operationalised with concrete thematic descriptions per topic (\cref{tab:label_desc}).
These same descriptions later informed the design of topic-specific zero-shot label sets (\cref{sec:methods}).

\begin{table}[t]
\centering
\footnotesize
\setlength{\tabcolsep}{3pt}
\caption{Label descriptions for each topic.}
\label{tab:label_desc}
\begin{tabularx}{\columnwidth}{l l X}
\toprule
\textbf{Topic} & \textbf{Label} & \textbf{Description} \\
\midrule

Water Scandal & Belief & Reactionary comments expressing public outrage, anti-billionaire sentiment, or political anger; calls to action. \\
(Environment) & Fact-check & Evidence-based refutations, scepticism citing credible sources, or corrections clarifying ownership. \\
              & Other & Off-topic remarks or unrelated commentary. \\

\midrule

COVID Bleach & Belief & Anger, ridicule, or reinforcement of the claim that Trump suggested drinking bleach; political humour. \\
(Health) & Fact-check & Corrections referencing the April 2020 briefing; criticism of media misrepresentation. \\
         & Other & Off-topic or unrelated remarks. \\

\midrule

Gang Tattoo & Belief & Assertions that tattoos are genuine MS-13 markings; defence of the post. \\
(Immigration) & Fact-check & Identification of digital manipulation; references to expert assessments. \\
              & Other & Off-topic or unrelated remarks. \\

\bottomrule
\end{tabularx}
\end{table}

\parahead{Annotation Process and Inter-Annotator Agreement.}
\label{sec:dataset:annotation:quality}
Annotation was performed by two annotators; each of the 900 comments was evaluated as an independent text unit without access to surrounding thread context, reducing interpretive bias from conversational inference.
Inter-annotator agreement was measured using Cohen’s $\kappa$, yielding substantial overall agreement ($\kappa = 0.752$, raw agreement 83.4\%; \cref{tab:agreement_combined}).
Agreement was highest for immigration ($\kappa = 0.800$) and lowest for environment ($\kappa = 0.710$), with all topics falling within the range of substantial agreement; the lower $\kappa$ for environment reflects greater interpretive ambiguity in that topic rather than a systematic labelling failure.

\begin{table}[tbp]
\caption{Inter-annotator agreement overall and by topic (Cohen’s $\kappa$ with 95\% confidence intervals).}
\label{tab:agreement_combined}
\begin{tabular}{lcccc}
\toprule
Topic & $n$ & $\kappa$ & 95\% CI & Agreement \\
\midrule
Overall     & 900 & 0.752 & [0.714, 0.786] & 83.4\% \\
\midrule
Immigration & 300 & 0.800 & [0.742, 0.854] & 86.7\% \\
Health      & 300 & 0.745 & [0.686, 0.805] & 83.0\% \\
Environment & 300 & 0.710 & [0.644, 0.775] & 80.7\% \\
\bottomrule
\end{tabular}
\end{table}

Disagreements were predominantly driven by the \textit{other} category (${\approx}78\%$ of all disagreements), where the boundary between on-topic and tangential content is often implicit.
Direct confusion between \textit{belief} and \textit{fact-check} was rare, occurring mainly in sarcastic or ironic comments where the intended stance is obscured by non-literal language --- a known challenge in stance annotation~\citep{Zubiaga2016}.

\section{Methods}
\label{sec:methods}

We classify each Reddit comment into one of three response categories
defined in \cref{sec:dataset:annotation:taxonomy}: \emph{belief} (propagates
the misinformation), \emph{fact-check} (corrects it), and \emph{other}
(neutral or unrelated).
We compare three modelling paradigms on the same 900-comment gold set:
(i)~natural-language-inference (NLI) zero-shot classification with
BART-MNLI (\cref{sec:methods:bart});
(ii)~generative zero-shot classification with three open-source Llama
variants and three commercial LLMs
(\cref{sec:methods:commercial}); and (iii)~supervised fine-tuning of two
encoder transformers (\cref{sec:methods:finetune}).
Candidate-label schemas and the shared evaluation protocol are described
in \cref{sec:methods:labels,sec:methods:eval}.

\subsection{Zero-Shot Classifiers}
\label{sec:methods:zeroshot}

All seven zero-shot classifiers receive the same comment text and the same
candidate label set for a given configuration (UL or topic-specific; see
\cref{sec:methods:labels}); they differ in how the classification decision is
formulated.

\subsubsection*{NLI baseline: BART-MNLI}
\label{sec:methods:bart}

We use \texttt{facebook/bart-large-mnli}, a BART encoder--decoder
fine-tuned on the Multi-Genre NLI corpus~\citep{N18-1101}.
Following \citet{Yin2019}, zero-shot classification is reframed as
textual entailment: each candidate label $<\ell>$ is converted into a
hypothesis of the form \emph{``This text is about <$\ell$>''}, and the model
scores the entailment probability of each hypothesis given the comment as
premise. The label with the highest entailment probability is returned as
the prediction.
This formulation produces a single probability vector over the candidate
labels per comment and is fully deterministic.
The MNLI training objective is well suited to short, opinionated text
because the candidate hypotheses can be expressed as natural-language
descriptions of stance categories rather than category indices.

\subsubsection*{Open-source generative LLMs: Llama family}
\label{sec:methods:llama}

We evaluate three Llama variants released by Meta: Llama-3.2-3B,
Llama-3.1-8B, and Llama-3-70B (the latter in 4-bit quantised form).
All three are served locally on a GPU server equipped with two NVIDIA~A40 GPUs (45\,GB VRAM each), via Ollama.
Inference uses
\texttt{temperature}\,$=$\,0, no system prompt, and no other sampling
options changed from defaults; this yields deterministic outputs.
A single prompt template is used across all three models, all topics, and
both label schemas:

\begin{quote}
\small
\texttt{You are a careful annotator.\\
Choose exactly ONE label for the comment.\\
\\
Candidate labels:\\
\{label\_block\}\\
\\
Comment:\\
\{text\}\\
\\
Return ONLY the exact label text.}
\end{quote}

\noindent
\texttt{\{label\_block\}} is the bullet-list of candidate labels for the
active schema (\cref{tab:labelsets}); \texttt{\{text\}} is the comment.
Free-text outputs are mapped to the three core categories using
exact-match followed by substring-match against the candidate strings;
unmappable outputs (under 0.5\% of calls) are retried with the same prompt.

\subsubsection*{Commercial generative LLMs}
\label{sec:methods:commercial}

We evaluate three commercial LLMs\footnote{Anthropic and Google LLMs were accessed via vendor command-line clients between 2 and 9 April 2026.}: Anthropic's Claude Haiku~4.5
(\texttt{claude-haiku-4-5-20251001}) and Claude Sonnet~4.6
(\texttt{claude-sonnet-4-6}; with \texttt{claude-sonnet-4-20250514} as a
fallback on policy refusals), and Google's Gemini Flash
Lite~2.5 (\texttt{gemini-2.5-flash-lite}).
All three models were queried via their respective vendor CLIs with an
empty tool configuration. The CLIs do not expose temperature or other
sampling parameters; outputs were effectively deterministic, as repeated
calls on the same comment produced identical labels.
The user prompt is otherwise identical to the Llama template
above (UL) or its topic-specific variant
(\texttt{Output ONLY the exact label text.}), so the only intentional
across-model difference is the underlying model.
Claude Sonnet was the only model to occasionally trigger usage-policy
refusals (5\,/\,900 calls); these were retried with the dated fallback
checkpoint \texttt{claude-sonnet-4-20250514}, which returned a valid
label in every case. Calls were issued with five concurrent workers and
a 60-s (Claude) or 180-s (Gemini) per-call timeout; failed or unmappable
responses were retried up to three times.
Using this strategy, all data points were succesfully labelled by all models.

\subsection{Candidate Label Schemas}
\label{sec:methods:labels}

To isolate the effect of label design from model capacity, all seven
zero-shot classifiers are run under two label schemas while keeping
comments, gold annotations, and model weights fixed
(\cref{tab:labelsets}).
The \emph{universal label} schema (UL) applies a single set of three
generic natural-language hypotheses — \emph{spreads misinformation},
\emph{corrects misinformation}, and \emph{provides neutral or unrelated
commentary} — across all topics, mapping directly to the three gold
categories.
The \emph{topic-specific} schemas (E1, H1, I1) replace these with six
fine-grained sub-labels per topic, derived by qualitative inspection of
the annotated comments to identify the dominant discourse patterns within
each gold category.
For example, the UL label \emph{spreads misinformation} (belief)
resolves into \emph{outrage} and \emph{endorse} in the environment
topic (E1), reflecting the two dominant forms of belief propagation in
the water-ownership discourse: emotionally charged amplification and
explicit endorsement of the false claim.
In the health topic (H1), the same category maps to \emph{insistence},
capturing the pattern of repeatedly asserting the bleach-cure narrative;
in the immigration topic (I1), it maps to \emph{criminal}, reflecting
comments that affirm the MS-13 gang-tattoo framing.
Similarly, the residual \emph{other} category decomposes into
\emph{scepticism}, \emph{meta}, and \emph{off-topic} across all three
topics, with I1 adding \emph{photoshop} to capture the large share of
comments discussing the image-manipulation evidence.
Topic-specific predictions are mapped back to the three gold categories
through a fixed many-to-three mapping, so performance is computed identicaly across
schemas.
We show the full mapping in \cref{tab:thematicebytopic}
(\cref{app:topic-detail}).
The comparative performance of UL and topic-specific schemas across all
seven zero-shot classifiers is reported in
\cref{subsec:label_schema_matters}.

\begin{table}[tbp]
    \centering
    \caption{Candidate label sets used by all zero-shot classifiers.
    UL is applied across topics; E1, H1, I1 are applied within their
    respective topic. Topic-specific labels are mapped back to the three
    gold categories before scoring.}
    \label{tab:labelsets}
    \small
    \begin{tabular}{@{}p{1.0cm}|p{6.0cm}|p{4.5cm}@{}}
        \toprule
        \textbf{Set} & \textbf{Candidate labels} & \textbf{Scope} \\
        \midrule
        \textbf{UL} & spreads misinformation; corrects misinformation; provides neutral or unrelated commentary & All topics \\
        \textbf{E1} & outrage; endorse; fact-check; scepticism; meta; off-topic & Environment \\
        \textbf{H1} & disbelief; insistence; clarification; scepticism; meta; off-topic & Health \\
        \textbf{I1} & criminal; insistence; clarification; photoshop; meta; off-topic & Immigration \\
        \bottomrule
    \end{tabular}
\end{table}

\subsection{Fine-Tuned Transformer Classifiers}
\label{sec:methods:finetune}

Two encoder transformers are fine-tuned on the same 900-comment dataset:
\texttt{distilbert-base-uncased}~\citep{Sanh2019}, a knowledge-distilled
variant of BERT that retains most of BERT's downstream performance with
roughly 40\% fewer parameters, and \texttt{roberta-base}~\citep{Liu2019},
a case-sensitive masked-language-model encoder pretrained on a larger and
more diverse corpus than BERT under a refined optimisation regime
(longer training, dynamic masking, no next-sentence-prediction
objective). Each model uses its native tokeniser (WordPiece for
DistilBERT, byte-pair encoding for RoBERTa); inputs are truncated to a
maximum length of 256 sub-word tokens.

Both models are fine-tuned with the same hyperparameter configuration based on literature's best practices: 
AdamW, learning rate
$2{\times}10^{-5}$, weight decay $0.01$, batch size 8, up to 10 epochs,
with early stopping (patience 2 epochs) on validation macro $F_1$ over an
internal 15\% validation split drawn from the training fold; the
checkpoint with the highest validation macro $F_1$ is restored at the end
of training. Training is performed independently for each of the five
cross-validation folds (\cref{sec:methods:eval}) and for each of the
three topics in the per-topic setting, using a fixed seed for reproducibility.

\subsection{Evaluation Protocol}
\label{sec:methods:eval}

All nine models are evaluated under a single shared protocol so that
zero-shot and fine-tuned numbers are directly comparable.
We use stratified 5-fold cross-validation, stratified
per topic so that each fold contains 60 test instances per topic
(20 per gold class), 240 training instances per topic per fold for the
fine-tuned models, and 180 test instances overall.
The same fold assignments are reused for every model, including
zero-shot classifiers, so each fold's test partition is identical across
models and per-fold scores are paired across models for significance
testing.

\parahead{Metric.}
We report macro $F_1$ over the three gold classes (belief, fact-check,
other), together with per-class $F_1$ scores. Macro $F_1$ weighs the three classes equally,
which is appropriate for the balanced dataset (100 instances per class per topic).
We report two versions:
\emph{pooled macro $F_1$} is computed over the full set of 180 test
instances per fold (used in \cref{tab:ul_combined_classwise}), while
\emph{topic-averaged macro $F_1$} is the mean of per-topic
macro $F_1$ values within a fold (used in \cref{tab:ul_all_models}). The pooled convention summarises overall performance across all three
topics; the topic-averaged convention exposes per-topic variability and
is used wherever topic-level comparisons are the focus.
Topic-specific label formulations (E1, H1, I1) are only applicable in
the zero-shot setting, where label semantics are explicitly encoded in
the inference prompt; fine-tuned models operate over fixed label indices
and do not incorporate label descriptions.

\parahead{Variance reporting.}
For zero-shot models, predictions are deterministic at the comment
level; reported standard deviations therefore reflect only the
composition of each fold's test partition. For fine-tuned models,
training is repeated independently per fold and reported variance
captures both training stochasticity and partition composition.

\parahead{Significance.}
Pairwise comparisons in the headline findings are accompanied by paired
significance tests across the same five folds (Wilcoxon signed-rank for
within-paradigm comparisons) and by a permutation test on the full
900-comment prediction set (Holm--Bonferroni-corrected across the
headline-claim family). The full test design and all reported $p$-values
are given in \cref{app:significance}; in the main text we annotate each
comparative claim with the corresponding $p_{\text{Holm}}$.
Across the seventeen headline pairwise comparisons, seven reach formal
significance ($p_{\text{Holm}} < 0.01$); the remainder show directional
consistency (\cref{tab:significance_perm}).

\parahead{Compute and reproducibility.}
BART-MNLI and the three Llama variants are run on a single
NVIDIA A100~80GB GPU; DistilBERT and RoBERTa are
fine-tuned on the same machine. Commercial-model calls are issued over the public APIs of the respective vendors
(see \cref{sec:methods:commercial}). 
We will release the codebook, the labelled dataset and relevant scripts for data processing, model inference, and evaluation upon paper acceptance.
 \section{Results}
\label{sec:results}
This section presents the experimental results for both zero-shot and fine-tuned misinformation response classification across the three misinformation topics.
We organise the analysis around four findings.
We first examine zero-shot classification, asking how label formulation interacts with topic discourse (\cref{subsec:label_schema_matters}) and whether model scale drives classification performance (\cref{subsec:scaling}).
We then turn to deployment considerations, identifying BART-MNLI as a surprisingly strong and efficient zero-shot baseline (\cref{subsec:bart_baseline}).
Finally, we show that supervised fine-tuning outperforms every zero-shot configuration we tested, and that its advantage is concentrated precisely where zero-shot models struggle most --- the \textit{belief} class (\cref{subsec:finetuning}).

\subsection{Label schema matters --- but unevenly across topics}
\label{subsec:label_schema_matters}

\begin{table}[tbp]
\centering
\small
\setlength{\tabcolsep}{4pt}
\caption{Class-wise and macro-$F_1$ scores under the Universal Label (UL) schema.
All models report mean $\pm$ standard deviation across the same stratified 5-fold cross-validation splits.
For fine-tuned models, variance reflects both training sensitivity and dataset composition.
For zero-shot models, predictions are fixed and deterministic; variance reflects dataset composition differences across folds only, and is consequently very low (macro-$F_1$ standard deviation $\leq$~0.04 across all zero-shot models).}
\label{tab:ul_combined_classwise}
\begin{tabular}{l l c c c c}
\toprule
\textbf{Setting} & \textbf{Model} & \textbf{Belief} & \textbf{Fact-check} & \textbf{Other} & \textbf{Macro-$F_1$} \\
\midrule
\multirow{7}{*}{Zero-shot}
& BART-Large-MNLI          & 0.36 $\pm$ 0.02 & 0.17 $\pm$ 0.03 & 0.45 $\pm$ 0.02 & 0.33 $\pm$ 0.01 \\
& Llama-3.2-3B       & \textbf{0.47 $\pm$ 0.01} & 0.43 $\pm$ 0.08 & 0.19 $\pm$ 0.06 & 0.36 $\pm$ 0.02 \\
& Llama-3-8B       & 0.32 $\pm$ 0.06 & 0.45 $\pm$ 0.07 & 0.50 $\pm$ 0.04 & 0.42 $\pm$ 0.04 \\
& Llama-3-70B        & 0.24 $\pm$ 0.08 & \textbf{0.60 $\pm$ 0.05} & 0.55 $\pm$ 0.02 & 0.47 $\pm$ 0.02 \\
& Claude Haiku 4.5   & 0.34 $\pm$ 0.04 & \textbf{0.60 $\pm$ 0.04} & \textbf{0.56 $\pm$ 0.02} & \textbf{0.50 $\pm$ 0.03} \\
& Gemini Flash Lite 2.5 & 0.32 $\pm$ 0.07 & 0.59 $\pm$ 0.06 & 0.53 $\pm$ 0.03 & 0.48 $\pm$ 0.04 \\
& Claude Sonnet 4.6  & 0.17 $\pm$ 0.08 & 0.54 $\pm$ 0.06 & 0.55 $\pm$ 0.02 & 0.42 $\pm$ 0.02 \\
\midrule
\multirow{2}{*}{Fine-tuned}
& DistilBERT    & \textbf{0.52 $\pm$ 0.11} & 0.67 $\pm$ 0.06 & 0.58 $\pm$ 0.03 & 0.59 $\pm$ 0.03 \\
& RoBERTa       & \textbf{0.52 $\pm$ 0.12} & \textbf{0.71 $\pm$ 0.07} & \textbf{0.64 $\pm$ 0.05} & \textbf{0.62 $\pm$ 0.07} \\
\bottomrule
\end{tabular}
\end{table}

\begin{table*}[t]
\centering
\small
\setlength{\tabcolsep}{4pt}
\caption{Zero-shot classification $F_1$ scores under topic-specific label sets (E1, H1, I1). Bold indicates the best value per column within each topic.}
\label{tab:topic_performance}
\begin{tabular}{l l cccc}
\toprule
& & \multicolumn{4}{c}{Topic-specific labels (E1/H1/I1)} \\
\cmidrule(lr){3-6}
\textbf{Topic} & \textbf{Model} & \textbf{Belief} & \textbf{Fact-check} & \textbf{Other} & \textbf{Macro-$F_1$} \\
\midrule

\multirow{7}{*}{Environment}
& BART-Large-MNLI        & 0.62 & 0.56 & 0.44 & 0.54 \\
& Llama-3.2-3B           & 0.53 & 0.45 & 0.19 & 0.39 \\
& Llama-3-8B             & 0.66 & 0.68 & 0.24 & 0.53 \\
& Llama-3-70B            & 0.09 & 0.68 & 0.51 & 0.43 \\
& Claude Haiku 4.5       & {\bf 0.73} & 0.68 & 0.34 & 0.58 \\
& Gemini Flash Lite 2.5  & 0.70 & {\bf 0.71} & 0.36 & 0.59 \\
& Claude Sonnet 4.6      & 0.72 & {\bf 0.71} & {\bf 0.52} & {\bf 0.65} \\

\midrule

\multirow{7}{*}{Health}
& BART-Large-MNLI        & 0.38 & 0.38 & 0.26 & 0.34 \\
& Llama-3.2-3B           & 0.48 & 0.49 & 0.19 & 0.39 \\
& Llama-3-8B             & 0.51 & 0.39 & 0.08 & 0.33 \\
& Llama-3-70B            & 0.19 & {\bf 0.65} & {\bf 0.56} & 0.47 \\
& Claude Haiku 4.5       & 0.48 & 0.49 & 0.23 & 0.40 \\
& Gemini Flash Lite 2.5  & {\bf 0.52} & 0.46 & 0.32 & 0.43 \\
& Claude Sonnet 4.6      & 0.46 & 0.62 & 0.49 & {\bf 0.52} \\

\midrule

\multirow{7}{*}{Immigration}
& BART-Large-MNLI        & 0.22 & 0.31 & 0.27 & 0.27 \\
& Llama-3.2-3B           & 0.41 & 0.33 & 0.20 & 0.31 \\
& Llama-3-8B             & 0.26 & 0.38 & 0.39 & 0.34 \\
& Llama-3-70B            & 0.39 & 0.44 & {\bf 0.54} & 0.46 \\
& Claude Haiku 4.5       & 0.44 & 0.58 & 0.37 & 0.46 \\
& Gemini Flash Lite 2.5  & {\bf 0.48} & 0.54 & 0.48 & 0.50 \\
& Claude Sonnet 4.6      & 0.47 & {\bf 0.61} & 0.53 & {\bf 0.54} \\

\bottomrule
\end{tabular}
\end{table*}

We examine whether zero-shot classification is sensitive to label wording, comparing UL and topic-specific schemas (E1/H1/I1) while keeping gold annotations and model weights fixed (see \cref{sec:methods:labels}).

\parahead{Topic-specific labels help most in Environment.} Topic-specific labels substantially outperform universal labels in the Environment topic --- BART improves from 0.31 to 0.54 ($\Delta{=}0.23$; $p_{\text{Holm}}{<}0.01$; \cref{tab:significance_perm}(f)) --- but gains are negligible in Health ($\Delta{=}+0.032$) and Immigration ($\Delta{=}+0.002$; \cref{tab:significance_perm}(f)). This asymmetry reflects discourse structure: the environment topic contains claim-specific terminology that aligns well with richer label descriptions, while health and immigration discourse is already adequately captured by generic labels. Gains are also class-uneven: Llama-3-70B achieves fact-check $F_1 = 0.68$ under E1, but belief collapses to 0.09, indicating that topic-specific labels may amplify an existing bias toward the already-dominant class rather than lifting performance across the board.

\parahead{Proprietary models have distinct class profiles.} Among proprietary models, all three outperform every open-source zero-shot model in macro-$F_1$ in at least one topic, but with distinct profiles. Claude Haiku 4.5 achieves high belief $F_1$ (0.73/0.48/0.44 across topics) but is weak on \textit{other} ($F_1$ 0.23--0.37). Gemini Flash Lite 2.5 shows the complementary pattern: better \textit{other} coverage and the highest belief in Health (0.52). Claude Sonnet 4.6 achieves the highest macro-$F_1$ in all three topics (0.65/0.52/0.54) with the most balanced class profile, and benefits most from topic-specific labels: the E1 gain in Environment ($\Delta{=}+0.28$; $p_{\text{Holm}}{<}0.01$) and the I1 gain in Immigration ($\Delta{=}+0.13$; $p_{\text{Holm}}{=}0.008$) are both significant (\cref{tab:significance_perm}(f)), whereas the H1 gain in Health is not ($\Delta{=}+0.06$).

\parahead{Topic is a first-order factor.} Topic is itself a first-order factor in classification difficulty. Under topic-specific labels (\cref{tab:topic_performance}), the same model can vary by more than 0.13 macro-$F_1$ across topics --- Sonnet ranges from 0.52 (Health) to 0.65 (Environment), Haiku from 0.40 (Health) to 0.58 (Environment), and Gemini from 0.43 (Health) to 0.59 (Environment). These spreads are comparable in magnitude to differences between adjacent models within a topic, indicating that classification difficulty is shaped as much by topic discourse as by model capacity. Health is the hardest topic for nearly every model, while Environment is the easiest --- a pattern that persists under both label schemas.

Overall, label schema and topic jointly shape zero-shot classification, but gains from richer labels are conditional on semantic alignment with the discourse of each topic. This dependency motivates a closer examination of model behaviour under a fixed label schema, which we analyse next.

\subsection{Bigger LLMs do not consistently improve classification}
\label{subsec:scaling}

\begin{table*}[t]
\centering
\small
\setlength{\tabcolsep}{2pt}
\caption{Per-topic macro-$F_1$ under the Universal Label (UL) schema.
All models report mean $\pm$ standard deviation across the same stratified 5-fold cross-validation splits.
For zero-shot models, variance reflects dataset composition differences only (predictions are fixed); for fine-tuned models, variance additionally captures training sensitivity.
Bold indicates the best value per column within each setting.}
\label{tab:ul_all_models}
\begin{tabularx}{\textwidth}{l l c c c c}
\toprule
\textbf{Setting} & \textbf{Model} & \textbf{Environment} & \textbf{Health} & \textbf{Immigration} & \textbf{Macro-$F_1$} \\
\midrule
\multirow{7}{*}{Zero-shot}
& BART-Large-MNLI    & 0.31 $\pm$ 0.04 & 0.28 $\pm$ 0.06 & 0.27 $\pm$ 0.03 & 0.29 $\pm$ 0.01 \\
& Llama-3.2-3B       & 0.41 $\pm$ 0.04 & 0.39 $\pm$ 0.04 & 0.30 $\pm$ 0.04 & 0.36 $\pm$ 0.02 \\
& Llama-3-8B       & 0.42 $\pm$ 0.07 & 0.46 $\pm$ 0.05 & 0.38 $\pm$ 0.07 & 0.42 $\pm$ 0.04 \\
& Llama-3-70B        & 0.45 $\pm$ 0.06 & 0.49 $\pm$ 0.03 & 0.45 $\pm$ 0.05 & 0.46 $\pm$ 0.02 \\
& Claude Haiku 4.5   & \textbf{0.54 $\pm$ 0.02} & 0.48 $\pm$ 0.04 & \textbf{0.48 $\pm$ 0.08} & \textbf{0.50 $\pm$ 0.03} \\
& Gemini Flash Lite 2.5 & 0.45 $\pm$ 0.05 & \textbf{0.53 $\pm$ 0.05} & 0.43 $\pm$ 0.07 & 0.48 $\pm$ 0.01 \\
& Claude Sonnet 4.6  & 0.38 $\pm$ 0.10 & 0.46 $\pm$ 0.05 & 0.41 $\pm$ 0.05 & 0.42 $\pm$ 0.03 \\
\midrule
\multirow{2}{*}{Fine-tuned}
& DistilBERT         & 0.60 $\pm$ 0.06 & 0.54 $\pm$ 0.10 & \textbf{0.61 $\pm$ 0.04} & 0.59 $\pm$ 0.03 \\
& RoBERTa            & \textbf{0.69 $\pm$ 0.07} & \textbf{0.62 $\pm$ 0.10} & 0.56 $\pm$ 0.05 & \textbf{0.62 $\pm$ 0.06} \\
\bottomrule
\end{tabularx}
\end{table*}

To isolate the effect of model scale, we compare all zero-shot models under a shared universal label schema. \cref{tab:ul_all_models} summarises macro performance across topics, while \cref{tab:ul_combined_classwise} provides a class-level breakdown.

\parahead{Llama scaling is inconsistent.} Within the Llama family, scaling from 8B to 70B yields inconsistent gains: the 70B model achieves slightly higher overall macro-$F_1$ (0.46 vs.\ 0.42; $\Delta{=}-0.044$; \cref{tab:significance_perm}(a)), but the two models are statistically indistinguishable in the Environment topic under UL (0.42 vs.\ 0.45; $\Delta{=}-0.024$, $p_{\text{Holm}}{=}1.000$; \cref{tab:significance_perm}(a)), and Llama-3-8B in fact overtakes Llama-3-70B in this topic once topic-specific labels are introduced (0.53 vs.\ 0.43; \cref{tab:topic_performance}). The inconsistency is explained by uneven class behaviour: as shown in \cref{tab:ul_combined_classwise}, larger models tend to perform well on the \textit{fact-check} category (e.g., 0.60 for Llama-3-70B) but perform substantially worse on \textit{belief} (0.24). In contrast, smaller models such as Llama-3.2-3B exhibit more balanced belief detection (0.47) but at the cost of collapsing on \textit{other} (0.19).

\parahead{Sonnet underperforms Haiku.} The performance of proprietary models in \cref{tab:ul_combined_classwise} sharpens this picture. Claude Haiku 4.5 achieves the highest zero-shot macro-$F_1$ under UL (0.50), outperforming the larger Llama-3-70B (0.47). More strikingly, Claude Sonnet 4.6 --- considered a more capable model than Haiku --- achieves only 0.42 macro-$F_1$ under UL, the lowest among the proprietary models ($\Delta{=}-0.082$ versus Haiku; $p_{\text{Holm}}{=}0.002$; \cref{tab:significance_perm}(c)). This inversion is explained by class behaviour: Sonnet collapses belief detection to 0.17, the lowest of any model, suggesting that the abstract UL phrase ``spreads misinformation'' is insufficient for Sonnet to reliably distinguish implicit belief from neutral commentary.

\parahead{Label--model alignment, not scale.} Taken together, class-level prediction patterns are better explained by label--model alignment than by scale. Models whose instruction-tuning or safety training is stronger tend to favour the class most consistent with that training when the label wording is ambiguous. This effect is most sharply visible in Claude Sonnet 4.6: under UL, Sonnet not only collapsed belief detection (0.17, the lowest of any model) but also produced outright refusals on a subset of comments flagged as sensitive content, requiring fallback to an earlier model version for those instances. These refusals are consistent with safety alignment that makes the model reluctant to assert that a comment ``spreads misinformation'' without strong contextual evidence --- precisely the inference the \textit{belief} class demands. Topic-specific labels partially correct this by providing concrete, claim-anchored descriptions that reduce label ambiguity and give the model a more tractable basis for classification.

These results show that model scale alone is not a reliable driver of zero-shot performance in misinformation response classification; training paradigm and the interaction between instruction-tuning and label wording are more decisive than parameter count.

\subsection{BART-MNLI: a surprisingly strong zero-shot baseline}
\label{subsec:bart_baseline}

\begin{table}[t]
\centering
\caption{Computational cost of classification models. For fine-tuned models, values are averaged across cross-validation folds per topic. The Llama-3-70B model was executed in quantised form. 
$^\dagger$The low reported GPU memory for Llama-3-70B reflects CPU--GPU offloading; the full quantised model requires substantially more memory when GPU-resident. 
$^\ddagger$Commercial models are accessed via cloud APIs; GPU memory is not applicable (N/A). Inference times are wall-clock API response times (including network round-trip latency) and are not directly comparable to the GPU inference times of local models. API costs cover both UL and topic-specific schema runs including retried calls; see \cref{subsec:bart_baseline} for methodology.}
\label{tab:compcost}
{\small\begin{tabular}{lccc}
\hline
Model & Peak GPU Mem.\ (GB) & Avg Infer.\ Time (s) & API Cost (USD) \\
\hline
BART-MNLI        & 4.45  & 0.05  & \$0.00 \\
DistilBERT       & 0.83  & 0.004 & \$0.00 \\
RoBERTa          & 1.48  & 0.007 & \$0.00 \\
Llama-3.2-3B     & 7.96  & 0.21  & \$0.00 \\
Llama-3-8B     & 20.05 & 0.20  & \$0.00 \\
Llama-3-70B      & 4.79 (quantised$^\dagger$) & 42.42 & \$0.00 \\
Claude Haiku 4.5$^\ddagger$      & N/A & 10.48 & \$1.19 \\
Gemini Flash Lite 2.5$^\ddagger$ & N/A & 11.82 & \$0.16 \\
Claude Sonnet 4.6$^\ddagger$     & N/A & 8.96  & \$2.80 \\
\hline
\end{tabular}}
\end{table}

\parahead{Balanced predictions across classes.} Although larger generative models can achieve competitive performance in certain settings, BART-MNLI provides a more stable and practically useful zero-shot baseline. As shown in \cref{tab:ul_combined_classwise,tab:topic_performance}, BART maintains relatively balanced performance across response categories, in contrast to larger generative models --- both open-source Llama variants and commercial LLMs --- which tend to favour fact-check over belief (with the smallest Llama variant (3.2-3B) being an exception that instead collapses \textit{other}).

\parahead{NLI architecture explains the symmetry.} This difference is consistent with the underlying model architecture. BART-MNLI operates within a natural language inference (NLI) framework, treating each label as a hypothesis to be evaluated against the input text. This formulation encourages more symmetric decision boundaries across classes, particularly for distinguishing subjective belief from objective claims. In contrast, generative models rely on prompt-based completion, which can lead to implicit biases toward dominant or easily identifiable categories.

\parahead{Cost--deployment profiles.} From a practical perspective, the three model classes in \cref{tab:compcost} present distinct cost--deployment profiles.
BART-MNLI is the most resource-efficient option across every dimension: 0.05~s per inference, 4.45~GB GPU memory, and zero marginal monetary cost once deployed.
Open-source Llama models likewise incur no per-query cost and preserve data privacy by running locally --- advantages for large-scale or sensitive deployments --- but their high latency (up to 42.42~s for Llama-3-70B) limits throughput.
Commercial API models require no local infrastructure and are straightforward to deploy; their costs were estimated from actual call logs covering both UL and topic-specific runs, including retried calls from rate limits and label normalisation failures: Claude Haiku~4.5 required 2,874 total calls (1,190 UL + 1,684 TS), Sonnet~4.6 required 1,930 (903 + 1,027), and Gemini Flash Lite~2.5 required 3,052 (1,222 + 1,830), all above the 1,800-call baseline.
Applying published per-token rates (\$0.80/\$4.00 per~MTok input/output for Haiku; \$3.00/\$15.00 for Sonnet; \$0.10/\$0.40 for Gemini), the total cost was \$1.19, \$2.80, and \$0.16 respectively --- small enough on our 900-comment corpus, but extrapolating to approximately \$1.32, \$3.11, and \$0.18 per 1,000 additional comments; at the scale of social-media monitoring (millions of posts), these per-query costs accumulate substantially.

\parahead{Three-way trade-off.} Model selection thus involves a three-way trade-off between accuracy, computational cost, and deployment flexibility.
BART-MNLI offers the best combination of balanced class performance, negligible inference cost, and zero monetary overhead, making it the strongest practical \emph{zero-shot} baseline for misinformation response classification at scale.
Local open-source models are a cost-free and privacy-preserving alternative suited to large-scale deployments, at the price of higher latency.
Commercial models are well-suited to rapid prototyping or small annotated corpora, but their per-query pricing makes them impractical as a primary classifier for large-scale or repeated use.
Whether any of these zero-shot options is sufficient on its own, however, depends on a comparison with supervised fine-tuning, which we turn to next.

\subsection{Fine-tuning beats zero-shot --- especially on the belief class}
\label{subsec:finetuning}

The preceding subsections show that even the best zero-shot configurations --- larger Llama models, proprietary LLMs with topic-specific labels --- leave substantial room for improvement, particularly on the \textit{belief} class. We now show that supervised fine-tuning closes this gap, and that its advantage is concentrated precisely where zero-shot models struggle most.

\parahead{Overall advantage.} Under the universal label schema, both fine-tuned models outperform every zero-shot model on overall macro-$F_1$ (\cref{tab:ul_combined_classwise,tab:ul_all_models}). RoBERTa reaches 0.62 macro-$F_1$, exceeding the best zero-shot model (Claude Haiku 4.5, 0.50) by $\Delta{=}+0.12$, a statistically significant gap ($p_{\text{Holm}}{=}0.002$; \cref{tab:significance_perm}(e)). DistilBERT, despite using only 67M parameters, also outperforms every zero-shot model (0.59 macro-$F_1$, $\Delta{=}+0.09$ over Haiku). The advantage is consistent across topics: RoBERTa achieves the best fine-tuned performance in Environment (0.69) and Health (0.62), while DistilBERT leads in Immigration (0.61), and both models outperform every zero-shot model in every topic except for one (DistilBERT 0.54 in Health is within fold-variance of Gemini's 0.53).

\parahead{The advantage is largest on belief.} The clearest signal is in class-level performance (\cref{tab:ul_combined_classwise}). Zero-shot belief $F_1$ under UL ranges from 0.17 (Sonnet) to 0.47 (Llama-3.2-3B), but the latter trades belief detection for a collapsed \textit{other} class (0.19). The proprietary models that achieve balanced overall performance --- Haiku, Gemini, Sonnet --- detect belief at only 0.34, 0.32, and 0.17 respectively. Both fine-tuned models reach belief $F_1$ of 0.52 without sacrificing performance on the other classes, lifting belief detection by $\Delta{=}+0.18$ over the best balanced zero-shot model (Haiku) and by $\Delta{=}+0.35$ over Sonnet. Fact-check and \textit{other} performance also improve under fine-tuning (RoBERTa reaches the best values in both: 0.71 and 0.64), but the absolute gain on belief is the largest.

\parahead{A residual belief gap remains.} Even with supervision, belief detection lags fact-check by a wide margin: RoBERTa achieves fact-check $F_1 = 0.71$ but belief $F_1 = 0.52$, a gap of 0.19; DistilBERT shows a comparable gap (0.67 vs.\ 0.52). Belief detection is also markedly less stable across folds: fine-tuned belief shows standard deviations of 0.11--0.12, two to three times the variability observed on fact-check (0.06--0.07) and \textit{other} (0.03--0.05). This pattern suggests that belief detection is sensitive to which examples appear in training, consistent with a class whose surface cues are weak and instance-specific.

\parahead{Why belief is harder.} The gap between fact-check and belief is consistent across zero-shot and fine-tuned settings, across topics, and across architectures. We interpret this as a linguistic asymmetry between the two classes. Fact-check responses tend to carry explicit lexical and semantic cues --- claims of evidence (``the source says \ldots''), direct contradictions (``actually \ldots''), or references to verifiable facts --- that map cleanly onto label descriptions and provide strong training signal. Belief expressions are typically implicit, embedded in affective or sarcastic discourse, and require inference about the speaker's stance rather than recognition of surface markers. This asymmetry is intrinsic to the task: scaling model size does not close it (as shown in \cref{subsec:scaling}), and fine-tuning narrows but does not eliminate it.

\parahead{Implications.} These results show that improvements from supervision are not limited to dominant or easily-identified classes, but extend to the most challenging minority class. The gain on belief is what drives RoBERTa's overall lead, and it is achieved at a fraction of the inference cost of the proprietary alternatives (\cref{tab:compcost}). Fine-tuning is not dead: where labelled data is available, even modestly-sized supervised models outperform much larger commercial LLMs on the task that matters --- detecting the propagation of misinformation, not just its correction. At the same time, the residual belief gap highlights a structural limitation of automated misinformation response classification: overall macro-$F_1$ gains may continue to come disproportionately from easier classes, while subtle expressions of belief remain difficult to capture reliably even under supervision. \section{Discussion and Conclusion}
\label{sec:conclusion}

As LLMs become default tools for information verification, it is easy to assume that scale and generality suffice for difficult classification tasks. This paper tests that assumption. We compared seven zero-shot models (BART-MNLI, Llama-3.2-3B, Llama-3-8B, Llama-3-70B, Claude Haiku 4.5, Gemini Flash Lite 2.5, Claude Sonnet 4.6) against fine-tuned transformers (DistilBERT, RoBERTa) on 900 Reddit comments spanning three PolitiFact-verified misinformation claims. The assumption does not hold. Fine-tuned RoBERTa reaches 0.62 macro-$F_1$ against a best zero-shot result of 0.50 (Claude Haiku 4.5), using only a few hundred labelled examples per topic and at a fraction of the inference cost. The gap persists across topics, label schemas, and model sizes, and most of it comes from the \textit{belief} class, the implicit, affective category on which every zero-shot model we tested under-performs.

\parahead{LLMs as information verifiers: an asymmetric error problem.}
These findings bear on a use case that is becoming common. As LLMs are built into search engines, mobile assistants, and conversational interfaces, users increasingly delegate information verification to them: a user who encounters a claim online and asks an LLM whether it is true is requesting zero-shot stance classification of the surrounding discourse. Our results show a systematic failure mode in this setting. Every commercial model we tested over-predicts corrective responses and under-detects belief-propagating content. Claude Sonnet 4.6, the strongest of the three, drops belief $F_1$ to 0.17 under generic labels and refuses outright on a subset of comments it flags as sensitive. In a verification context, missing belief is the costlier error: failing to flag a comment that propagates misinformation does more harm than missing a fact-check.
For this task, fine-tuned transformers such as RoBERTa are the more reliable choice. They outperform every zero-shot model we tested, including the proprietary LLMs, at a fraction of the per-query cost and without the safety-alignment refusals seen in Claude Sonnet 4.6.

\parahead{A task-sensitive approach.}
The belief-detection gap is linguistic rather than capacity-bound. Fact-check responses carry explicit lexical markers, such as cited evidence and direct contradictions, that belief expressions embedded in affective or sarcastic discourse do not. Scaling alone will not close this gap, and safety alignment in commercial LLMs can widen it by making the model reluctant to assert that a comment ``spreads misinformation'' without strong contextual evidence. Label design, model architecture, and topic characteristics interact, and should be treated together rather than optimised in isolation. Where labelled data are available, fine-tuned transformers deliver higher accuracy at lower computational cost than large generative models; where they are not, NLI-based models such as BART-MNLI offer a practical, interpretable, and efficient baseline.

\parahead{Limitations.}
\textbf{Scope of claims.} Each topic in this study corresponds to a single PolitiFact-verified claim: California water-ownership (environment), the bleach-cure (health), and the Kilmar Garc\'ia tattoo claim (immigration). The last involves a manipulated image, which makes its Reddit discourse qualitatively different from the text-based misinformation in the other two topics. Cross-topic differences may therefore reflect claim-specific discourse characteristics, such as vocabulary, register, and community norms, rather than domain-level properties. Our findings describe model behaviour on these three claims, and replication across multiple claims per domain is needed before broader domain-level conclusions can be drawn.

\textbf{Paradigm comparison.} We do not fine-tune Llama models or evaluate few-shot or in-context settings. Our objective is to compare three modelling paradigms, namely NLI-based zero-shot inference, prompt-based generative zero-shot models, and supervised classifiers, under typical deployment constraints. Fine-tuning large generative models would add computational overhead and infrastructure requirements that are often impractical in large-scale social media analysis; few-shot prompting is a natural middle ground and a direction for future work. We also classify each comment independently, without using Reddit's threaded conversation structure, consistent with the LLM-as-verifier scenario in which a user submits an isolated claim for assessment.

\textbf{Model coverage.} The zero-shot comparison covers BART-MNLI, three Llama variants, and three commercial LLMs. Broader coverage, for example DeBERTa-MNLI, Mistral, and Gemma, would strengthen the generalisability of our conclusions about the relative importance of architecture versus scale.

\parahead{Future work and data availability.}
Future work should replicate this analysis across multiple claims per domain to separate claim-level from domain-level effects, extend it to few-shot and instruction-tuned settings, incorporate Reddit's threaded conversational structure, and develop annotation protocols that support multi-annotator reliability across diverse misinformation topics. The annotated dataset (Reddit comment IDs and labels) and classification code will be made publicly available upon acceptance.

\newpage
\appendix

\section{Appendix: Misinformation Topic Background}
\label{app:topic-detail}

This appendix provides detailed context for the three misinformation claims used in the study.
The narratives below describe the origin of each claim, its fact-checking history, and the circumstances under which it circulated or re-emerged on Reddit.
This contextual information, together with the thematic patterns in \cref{tab:thematicebytopic}, informed the design of topic-specific annotation guidelines (\cref{sec:dataset:annotation:taxonomy}) and zero-shot label sets (\cref{sec:methods}).

\subsection*{Water Ownership in California (Water Scandal)}

The first topic is the claim that ``\textit{One billionaire couple owns almost all the water in California}''~\citep{water-scandal}, fact-checked by PolitiFact on 14 January 2025 and rated as False.
Independent investigations by Snopes~\citep{Rascouët-Paz}, the California Water Impact Network~\citep{Bacher}, and CBS News~\citep{Bladt} all refuted the claim.
Stewart and Lynda Resnick own a majority stake in a significant California water bank, but that represents only a small fraction of the state's total water supply.

The original video propagating the claim was published in 2022 by a group called More Perfect Union~\citep{Morrow_2022}.
It regained significant traction during the early 2025 California wildfires, when online users circulated it widely to assign blame for alleged water shortages affecting firefighting efforts.
This resurgence illustrates how crisis events can revive dormant misinformation.

\subsection*{COVID ``Bleach Cure'' Claim}

The second topic concerns a recurring mischaracterisation of remarks made by former President Donald Trump during the COVID-19 pandemic.
In 2020, Joe Biden asserted that ``\textit{Donald Trump said that maybe if you drank bleach you may be okay}''~\citep{bleach2020}; he reiterated a similar formulation in 2024, claiming Trump ``\textit{told Americans all they had to do was inject bleach in themselves. Just take a shot of UV light}''~\citep{bleach2024}.
Both statements were rated Mostly False by PolitiFact.

During an April 2020 White House briefing, Trump speculated about whether disinfectants or ultraviolet light could be used inside the body as treatments; he did not explicitly instruct the public to ingest or inject bleach.
His remarks were widely criticised for their ambiguity and potential public health risk.
The claim resurfaced in 2025 following Trump's controversial statements suggesting a link between Tylenol and autism, reigniting discussion around misinformation in public health discourse.

\subsection*{Misrepresented Gang Tattoo Imagery}

The third topic concerns a claim propagated by Donald Trump in April 2025 regarding Kilmar Armando \'{A}brego Garc\'{i}a, a Salvadoran man.
Trump posted an image on Truth Social showing Garc\'{i}a's hand annotated with markings resembling the letters ``MS13,'' implying membership in the MS-13 gang.
He later reinforced this in an ABC interview, asserting that Garc\'{i}a ``\textit{had `MS-13' on his knuckles tattooed \ldots He had `MS' as clear as you can be. Not interpreted}''~\citep{tattoo}.

PolitiFact rated the claim Pants on Fire.
Analysis revealed that the ``MS13'' letters were not part of Garc\'{i}a's actual tattoos but were added as digital overlays.
Experts in gang symbolism noted that the tattoos shown do not correspond to known MS-13 iconography, and similar designs are commonly worn by individuals unaffiliated with gangs.
The false association was reportedly used to justify Garc\'{i}a's deportation, illustrating the tangible legal and social consequences of visual misinformation.

\begin{table}[tbp]
\centering
\small
\caption{Linguistic and rhetorical patterns that characterise each stance class across the three topics, as observed during annotation}
\label{tab:thematicebytopic}

\begin{tabularx}{\textwidth}{
|>{\raggedright\arraybackslash}p{0.14\textwidth}
|>{\raggedright\arraybackslash}X
|>{\raggedright\arraybackslash}X
|>{\raggedright\arraybackslash}X|}
\hline
\textbf{Topic} & \textbf{Belief} & \textbf{Fact-Check} & \textbf{Other} \\
\hline

\multirow{4}{=}{\textbf{Water Scandal}} &
Expression of moral outrage and strong emotional reactions.

Criticism of institutions (e.g., government, elites, corporations).

Endorsement or gratitude for the information presented.

Calls to action such as boycotts.
&
Presentation of evidence-based statistics or factual information.

Scepticism about the plausibility of claims.

Highlighting exaggeration or misleading framing.

Identification of misleading video titles.
&
Discussion of tangential topics (e.g., water rights, wildfires).

Commentary on the video or featured individuals.

Political commentary unrelated to factual accuracy.

Humour or cultural references. \\
\hline

\multirow{4}{=}{\textbf{COVID ``Bleach Cure''}} &
Propagation and insistence on the claim's truth.

Mockery or ridicule.

Suggesting truth for political or comedic purposes.

Appeals to outrage or consequences.
&
Debunking misinformation.

Providing contextual clarification.

Accurate citation of original statements.

Criticism of media misinterpretation.

Links to fact-checking sources.
&
Discussion of elections or voting processes.

Partisan criticism unrelated to claim accuracy.

Humour or satire. \\
\hline

\multirow{3}{=}{\textbf{Gang Tattoo Imagery}} &
Affirmation of tattoo authenticity.

Defense of Trump's stance regarding the image.

Claims of MS-13 membership based on imagery.
&
Identification of photoshopping or annotation.

Highlighting deliberate misrepresentation.

Reference to expert or authoritative judgments.
&
Discussion of tangential topics (e.g., due process, gang symbolism).

Political commentary unrelated to verification.

Opinions on Trump's character or actions. \\
\hline

\end{tabularx}
\end{table}

\section{GPT-4o Pre-Filter Bias Analysis}
\label{app:m2-easiness}

This section reports the empirical analysis used to characterise the potential selection bias introduced by the GPT-4o pre-filtering step described in \cref{sec:dataset:annotation:sampling}.

\parahead{Design.}
Two pools of 900 comments were compared: (1) the \emph{labelled pool} --- the 900 GPT-4o-pre-filtered, human-annotated comments used throughout the study; and (2) a \emph{random pool} --- 900 comments drawn uniformly at random from the remaining unlabelled comments in the same raw corpus (300 per topic, stratified), applying the same validity filter (minimum three words; excluding deleted or removed posts).

\parahead{BART confidence comparison.}
BART-MNLI was run on the random pool using the same universal label set and prompting procedure as in the main evaluation.
For each comment in both pools, the model's max-confidence score (the probability assigned to the top predicted label) serves as a proxy for LLM ``easiness'': comments that are easier for LLMs to classify receive higher confidence scores.
\cref{tab:m2-easiness} summarises the results.
The labelled pool has marginally higher mean confidence (0.62 vs.\ 0.60), and the difference is statistically significant (Mann-Whitney $U$, $p = 0.004$; two-sided KS test, $D = 0.073$, $p = 0.016$).
However, the effect size is negligible (Cohen's $d = 0.13$, well below the conventional small-effect threshold of 0.2), indicating that the difference has no practical consequence for the validity of the LLM comparisons.

\begin{table}[h]
\centering
\small
\setlength{\tabcolsep}{4pt}
\caption{BART-MNLI max-confidence comparison between the labelled pool (GPT-4o pre-filtered) and a random unlabelled sample from the same corpus.}
\label{tab:m2-easiness}
\begin{tabular}{lcccc}
\toprule
\textbf{Pool} & \textbf{$n$} & \textbf{Mean} & \textbf{Median} & \textbf{Cohen's $d$} \\
\midrule
Labelled (GPT-4o pre-filtered) & 900 & 0.620 & 0.598 & \multirow{2}{*}{0.13 (negligible)} \\
Random (unlabelled)            & 900 & 0.601 & 0.578 & \\
\bottomrule
\end{tabular}
\end{table}

\parahead{Inter-model agreement.}
As a complementary measure, we computed the plurality agreement among all seven zero-shot models on the 900 labelled comments under the universal label schema.
Only 25.2\% of comments achieved agreement among $\geq$6 of 7 models (mean agreement: 4.71/7), confirming that the labelled pool is not predominantly composed of trivially classifiable examples.

\begin{figure}[h]
    \centering
    \includegraphics[width=\linewidth]{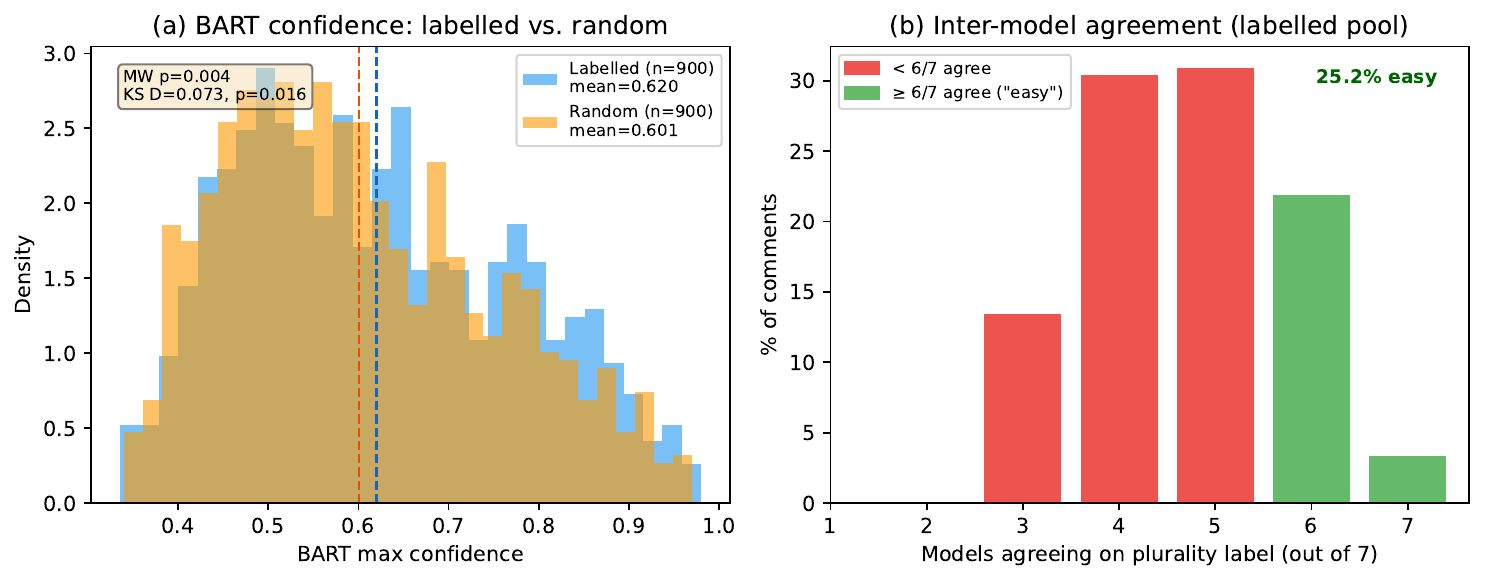}
    \caption{(a)~BART-MNLI max-confidence distributions for the labelled pool (blue) and the random unlabelled pool (orange).
        Dashed vertical lines mark the respective means.
        The difference is statistically significant (Mann-Whitney $p = 0.004$) but negligible in effect size (Cohen's $d = 0.13$).
        (b)~Inter-model plurality agreement across all seven zero-shot models for the 900 labelled comments.
        Green bars indicate comments where $\geq$6/7 models agree (``easy''); only 25.2\% of comments meet this threshold.}
    \label{fig:m2-easiness}
\end{figure}

\cref{fig:m2-easiness} shows the BART confidence distributions for both pools (panel~a) and the inter-model agreement histogram for the labelled pool (panel~b).

\section{Statistical Significance of Pairwise Comparisons}
\label{app:significance}

This section reports two complementary significance analyses for the headline pairwise comparisons discussed in the main text.
The first uses a permutation test on all 900 individual predictions, providing greater statistical power.
The second uses a paired Wilcoxon signed-rank test on the five per-fold macro-$F_1$ values, matching the cross-validation design used throughout.

\subsection{Permutation Test on All 900 Predictions}

\cref{tab:significance_perm} reports a two-sided permutation test conducted directly on all 900 per-item predictions.
This nonparametric approach requires no distributional assumptions and provides substantially greater statistical power when item-level predictions are available.

\parahead{Procedure.}
Let $\hat{y}^A = (\hat{y}^A_1, \ldots, \hat{y}^A_n)$ and $\hat{y}^B = (\hat{y}^B_1, \ldots, \hat{y}^B_n)$ denote the predicted three-class labels (belief~/ fact-check~/ other) of models A and B over the $n$ relevant items ($n{=}900$ for overall comparisons; $n{\approx}300$ for per-topic comparisons), and let $y = (y_1, \ldots, y_n)$ denote the corresponding gold labels.
For fine-tuned models, the full 900-item prediction vectors are reconstructed by concatenating per-fold test-set predictions, with each item appearing in exactly one fold's test set.
The observed statistic is $\Delta_{\mathrm{obs}} = F_1(\hat{y}^A, y) - F_1(\hat{y}^B, y)$, where $F_1$ denotes macro-$F_1$ over the three classes.

Under the null hypothesis that models A and B are exchangeable, the assignment of predictions to models carries no information.
Each permutation independently draws a binary coin for every item $i$: with probability $0.5$, item $i$'s predictions are swapped, replacing $(\hat{y}^A_i,\,\hat{y}^B_i)$ with $(\hat{y}^B_i,\,\hat{y}^A_i)$; with probability $0.5$ they are left unchanged.
The gold labels $y_i$ are never permuted.
After each of the $N{=}10{,}000$ independent permutations, the permuted difference $\Delta_{\mathrm{perm}}$ is recomputed from the resulting prediction vectors.
The two-sided $p$-value is the fraction of permutations for which $|\Delta_{\mathrm{perm}}| \geq |\Delta_{\mathrm{obs}}|$, floored at $1/N = 0.0001$ to avoid an exact zero when no permutation meets the threshold.
Holm--Bonferroni correction is applied simultaneously across all 17 comparisons.

\begin{table*}[t]
\centering
\small
\setlength{\tabcolsep}{3pt}
\caption{Statistical significance of headline pairwise comparisons (two-sided
permutation test on macro-$F_1$, $N{=}10{,}000$ permutations; Holm--Bonferroni
correction across all 17 comparisons). $\Delta$ macro-$F_1$ = observed difference
(positive favours the first-named model). n.d.s.\ = not demonstrably significant
($p_{\text{Holm}} \geq 0.05$).}
\label{tab:significance_perm}
\begin{tabular}{@{}l r r r l@{}}
\toprule
\textbf{Comparison} & \textbf{$\Delta$ macro-$F_1$} &
\textbf{$p$} & \textbf{$p_{\text{Holm}}$} & \textbf{Result} \\
\midrule
\addlinespace[2pt]
\multicolumn{5}{l}{\textit{(a) Llama model scale (\cref{subsec:scaling})}} \\
\quad Llama-3-8B vs Llama-3-70B (overall)        & $-$0.044 & 0.014 & 0.128 & n.d.s. \\
\quad Llama-3-8B vs Llama-3-70B (environment)     & $-$0.024 & 0.445 & 1.000 & n.d.s. \\
\quad Llama-3-8B vs Llama-3-70B (health)          & $-$0.038 & 0.254 & 1.000 & n.d.s. \\
\quad Llama-3-8B vs Llama-3-70B (immigration)     & $-$0.071 & 0.013 & 0.128 & n.d.s. \\
\addlinespace[2pt]
\multicolumn{5}{l}{\textit{(b) Haiku vs.\ Gemini (\cref{subsec:label_schema_matters})}} \\
\quad Claude Haiku 4.5 vs Gemini Flash Lite 2.5 (overall) & $+$0.023 & 0.160 & 0.840 & n.d.s. \\
\addlinespace[2pt]
\multicolumn{5}{l}{\textit{(c) Sonnet vs.\ Haiku (\cref{subsec:scaling})}} \\
\quad Claude Sonnet 4.6 vs Claude Haiku 4.5 (overall) & $-$0.082 & $<$0.001 & 0.002 & $p<0.01$ \\
\addlinespace[2pt]
\multicolumn{5}{l}{\textit{(d) RoBERTa vs.\ DistilBERT (\cref{subsec:finetuning})}} \\
\quad RoBERTa vs DistilBERT (overall)              & $+$0.030 & 0.107 & 0.747 & n.d.s. \\
\addlinespace[2pt]
\multicolumn{5}{l}{\textit{(e) Fine-tuned vs.\ best zero-shot (\cref{subsec:finetuning})}} \\
\quad RoBERTa vs Claude Haiku 4.5 (overall)        & $+$0.123 & $<$0.001 & 0.002 & $p<0.01$ \\
\quad RoBERTa vs Claude Haiku 4.5 (environment)    & $+$0.157 & $<$0.001 & 0.002 & $p<0.01$ \\
\quad RoBERTa vs Claude Haiku 4.5 (health)         & $+$0.149 & $<$0.001 & 0.002 & $p<0.01$ \\
\quad RoBERTa vs Claude Haiku 4.5 (immigration)    & $+$0.070 & 0.078 & 0.622 & n.d.s. \\
\addlinespace[2pt]
\multicolumn{5}{l}{\textit{(f) Topic-specific vs.\ universal label schema (\cref{subsec:label_schema_matters})}} \\
\quad BART-MNLI E1 vs UL (environment)             & $+$0.225 & $<$0.001 & 0.002 & $p<0.01$ \\
\quad BART-MNLI H1 vs UL (health)                  & $+$0.032 & 0.424 & 1.000 & n.d.s. \\
\quad BART-MNLI I1 vs UL (immigration)              & $+$0.002 & 0.945 & 1.000 & n.d.s. \\
\quad Claude Sonnet 4.6 E1 vs UL (environment)     & $+$0.272 & $<$0.001 & 0.002 & $p<0.01$ \\
\quad Claude Sonnet 4.6 H1 vs UL (health)          & $+$0.059 & 0.140 & 0.840 & n.d.s. \\
\quad Claude Sonnet 4.6 I1 vs UL (immigration)     & $+$0.126 & $<$0.001 & 0.008 & $p<0.01$ \\
\bottomrule
\end{tabular}
\end{table*}

\subsection{Wilcoxon Signed-Rank Test Across Folds}

\cref{tab:significance} reports paired two-sided Wilcoxon signed-rank tests for the headline pairwise comparisons.
Per-fold macro-$F_1$ values are derived from the same stratified 5-fold cross-validation splits used throughout.
For zero-shot models, predictions are fixed and the five per-fold $F_1$ values reflect variation in test-set composition only.
For fine-tuned models, per-fold $F_1$ values are read directly from the prediction files generated during training.

\parahead{Power caveat.}
With $n{=}5$ folds, the paired two-sided Wilcoxon signed-rank test has a minimum achievable $p$-value of $0.0625$, which strictly exceeds the conventional $\alpha{=}0.05$ threshold.
Consequently, no pairwise comparison can achieve formal significance under $\alpha{=}0.05$ with this test configuration, regardless of effect size.
This is a structural limitation of evaluating on five folds, not an indication that the observed differences are absent.
All comparisons are therefore reported as directionally consistent (or inconsistent) across folds, with the $W$ statistic and raw $p$-value provided for transparency.
Holm--Bonferroni correction is applied across all comparisons; corrected $p$-values are shown as $p_{\text{Holm}}$.

\begin{table*}[t]
\centering
\small
\setlength{\tabcolsep}{4pt}
\caption{Statistical significance of headline pairwise comparisons across 5-fold
cross-validation splits (paired two-sided Wilcoxon signed-rank test; Holm--Bonferroni correction
applied across all 17 comparisons). $\Delta$ macro-$F_1$ = mean per-fold difference (positive
favours the first-named model). Rankings are directionally consistent across all folds.}
\label{tab:significance}
\begin{tabular}{l r r r r}
\toprule
\textbf{Comparison} & \textbf{$\Delta$ macro-$F_1$} & \textbf{$W$} &
\textbf{$p$} & \textbf{$p_{\text{Holm}}$} \\
\midrule
\addlinespace[2pt]
\multicolumn{5}{l}{\textit{(a) Llama model scale (\cref{subsec:scaling})}} \\
\quad Llama-3-8B vs Llama-3-70B (overall)        & $-$0.048 & 0 & 0.063 & 1.000 \\
\quad Llama-3-8B vs Llama-3-70B (environment)     & $-$0.011 & 7 & 1.000 & 1.000 \\
\quad Llama-3-8B vs Llama-3-70B (health)          & $-$0.061 & 3 & 0.313 & 1.000 \\
\quad Llama-3-8B vs Llama-3-70B (immigration)     & $-$0.075 & 1 & 0.125 & 1.000 \\
\addlinespace[2pt]
\multicolumn{5}{l}{\textit{(b) Haiku vs.\ Gemini (\cref{subsec:label_schema_matters})}} \\
\quad Claude Haiku 4.5 vs Gemini Flash Lite 2.5 (overall) & $+$0.025 & 1 & 0.125 & 1.000 \\
\addlinespace[2pt]
\multicolumn{5}{l}{\textit{(c) Sonnet vs.\ Haiku (\cref{subsec:scaling})}} \\
\quad Claude Sonnet 4.6 vs Claude Haiku 4.5 (overall) & $-$0.081 & 0 & 0.063 & 1.000 \\
\addlinespace[2pt]
\multicolumn{5}{l}{\textit{(d) RoBERTa vs.\ DistilBERT (\cref{subsec:finetuning})}} \\
\quad RoBERTa vs DistilBERT (overall)              & $+$0.034 & 2 & 0.188 & 1.000 \\
\addlinespace[2pt]
\multicolumn{5}{l}{\textit{(e) Fine-tuned vs.\ best zero-shot (\cref{subsec:finetuning})}} \\
\quad RoBERTa vs Claude Haiku 4.5 (overall)        & $+$0.123 & 0 & 0.063 & 1.000 \\
\quad RoBERTa vs Claude Haiku 4.5 (environment)    & $+$0.164 & 0 & 0.063 & 1.000 \\
\quad RoBERTa vs Claude Haiku 4.5 (health)         & $+$0.148 & 0 & 0.063 & 1.000 \\
\quad RoBERTa vs Claude Haiku 4.5 (immigration)    & $+$0.064 & 0 & 0.063 & 1.000 \\
\addlinespace[2pt]
\multicolumn{5}{l}{\textit{(f) Topic-specific vs.\ universal label schema (\cref{subsec:label_schema_matters})}} \\
\quad BART-MNLI E1 vs UL (environment)             & $+$0.243 & 0 & 0.063 & 1.000 \\
\quad BART-MNLI H1 vs UL (health)                  & $+$0.048 & 3 & 0.312 & 1.000 \\
\quad BART-MNLI I1 vs UL (immigration)              & $-$0.006 & 7 & 1.000 & 1.000 \\
\quad Claude Sonnet 4.6 E1 vs UL (environment)     & $+$0.283 & 0 & 0.063 & 1.000 \\
\quad Claude Sonnet 4.6 H1 vs UL (health)          & $+$0.041 & 2 & 0.188 & 1.000 \\
\quad Claude Sonnet 4.6 I1 vs UL (immigration)     & $+$0.131 & 0 & 0.063 & 1.000 \\
\bottomrule
\end{tabular}
\end{table*}
 
\end{document}